\documentclass[letterpaper, 10 pt, conference]{ieeeconf}  

\IEEEoverridecommandlockouts                              

\usepackage{float}
\usepackage[caption = true]{subfig}
\usepackage{url}
\usepackage{bm,bbold}
\usepackage{textcomp}
\usepackage{indentfirst}
\usepackage{xcolor}

\let\labelindent\relax
\usepackage{enumitem}
\usepackage{graphicx}
\usepackage{amsmath}
\usepackage{amssymb}
\usepackage[draft,breaklinks=true,bookmarks=false]{hyperref}
\usepackage{booktabs}

\graphicspath{{figure/}}

\newcommand{\argmin}{\mathop{\rm argmin}}

\newcommand{\minimize}{\mathop{\rm minimize}}
\newcommand{\subjectto}{\mathop{\rm subject\:to}}

\title{\LARGE \bf
PedX: Benchmark Dataset for Metric 3D Pose Estimation of Pedestrians in Complex Urban Intersections
}

\author{
    Wonhui Kim$^{1}$, Manikandasriram Srinivasan Ramanagopal$^{1}$, Charles Barto$^{1}$, Ming-Yuan Yu$^{1}$,\\Karl Rosaen$^{1}$, Nick Goumas$^{1}$, Ram Vasudevan$^{2}$ and Matthew Johnson-Roberson$^{3}$
    \thanks{*This work was supported by a grant from Ford Motor Company via the Ford-UM Alliance under award N022884}
    \thanks{$^{1}$ Wonhui Kim, Manikandasriram S. R., Charles Barto, Ming-Yuan Yu, Karl Rosaen, and Nick Goumas are research assistants and engineers with UM and Ford Center for Autonomous Vehicles Laboratory, University of Michigan,
        Ann Arbor, MI 48105, USA
        {\tt\small https://fcav.engin.umich.edu/}}%
    \thanks{$^{2}$Ram Vasudevan is with Faculty of Mechanical Engineering,
        University of Michigan, Ann Arbor, MI 48105, USA
        {\tt\small ramv@umich.edu}}%
    \thanks{$^{3}$Matthew Johnson-Roberson is with Faculty of Naval Architecture \& Marine Engineering, Electrical Engineering and Computer Science,
        University of Michigan, Ann Arbor, MI 48105, USA
        {\tt\small mattjr@umich.edu}}%
}

\begin{document}

\maketitle
\thispagestyle{empty}
\pagestyle{empty}

\begin{abstract}
This paper presents a novel dataset titled \textit{PedX}, a large-scale multimodal collection of pedestrians at complex urban intersections. \textit{PedX} consists of more than 5,000 pairs of high-resolution (12MP) stereo images and LiDAR data along with providing 2D and 3D labels of pedestrians.
We also present a novel 3D model fitting algorithm for automatic 3D labeling harnessing constraints across different modalities and novel shape and temporal priors. All annotated 3D pedestrians are localized into the real-world metric space, and the generated 3D models are validated using a mocap system configured in a controlled outdoor environment to simulate pedestrians in urban intersections. We also show that the manual 2D labels can be replaced by state-of-the-art automated labeling approaches, thereby facilitating automatic generation of large scale datasets.
\end{abstract}

\section{Introduction}
\label{sec:introduction}
Driving in complex urban environments is one of the major challenges for autonomous vehicles (AVs). For AVs to operate in an environment crowded with people, understanding pedestrian pose, motion, behavior, and intention will greatly increase our ability to function safely and efficiently.

In computer vision, estimating human pose has been a long standing problem. The recent application of deep neural networks has generated state-of-the-art results for 2D body pose estimation \cite{cao2017realtime}, which has inspired extensions to the 3D pose estimation \cite{mehta2017vnect, zhou2017towards, zhou2016sparseness, chen20163d, martinez_2017_3dbaseline}. However, gathering ground truth 3D pose data is challenging. Motion capture (mocap) systems have been the primary generator of ground truth 3D pose data, but have restricted the variety and complexity of the 3D scenes that can be captured~\cite{sigal2010humaneva, h36m_pami}. For example, with mocap systems it is difficult to capture naturalistic in-the-wild scenes with groups of people who are moving and interacting. To overcome those technical limitations, this paper develops both a dataset and a ground truth generation approach to facilitate generating 3D poses on in-the-wild images without relying on mocap.

\begin{figure}[!t]
    \centering
    \includegraphics[width=0.48\linewidth]{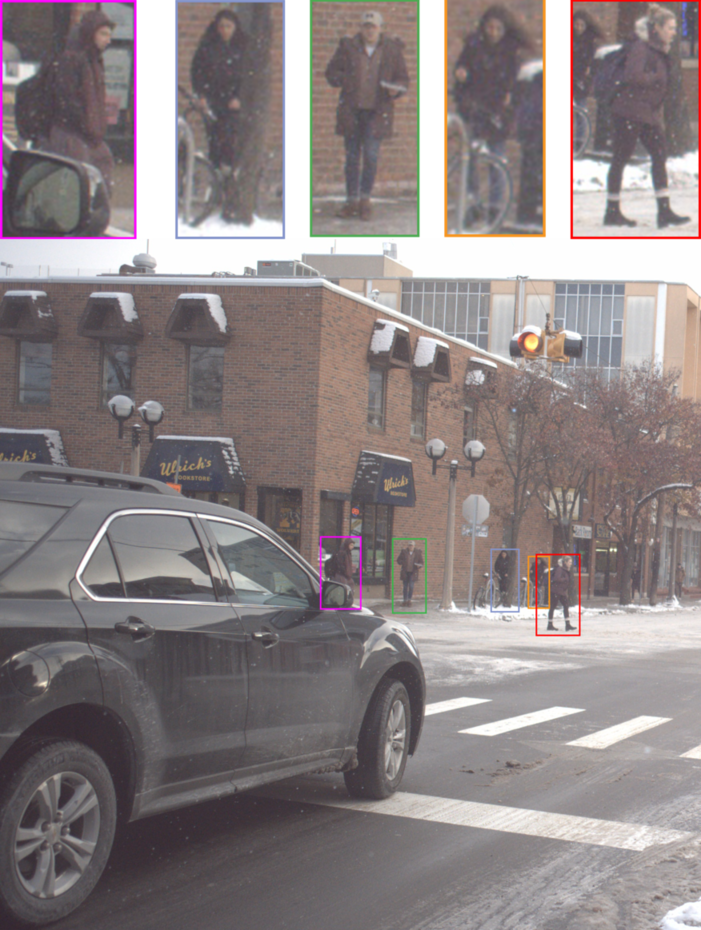}
    \includegraphics[width=0.48\linewidth]{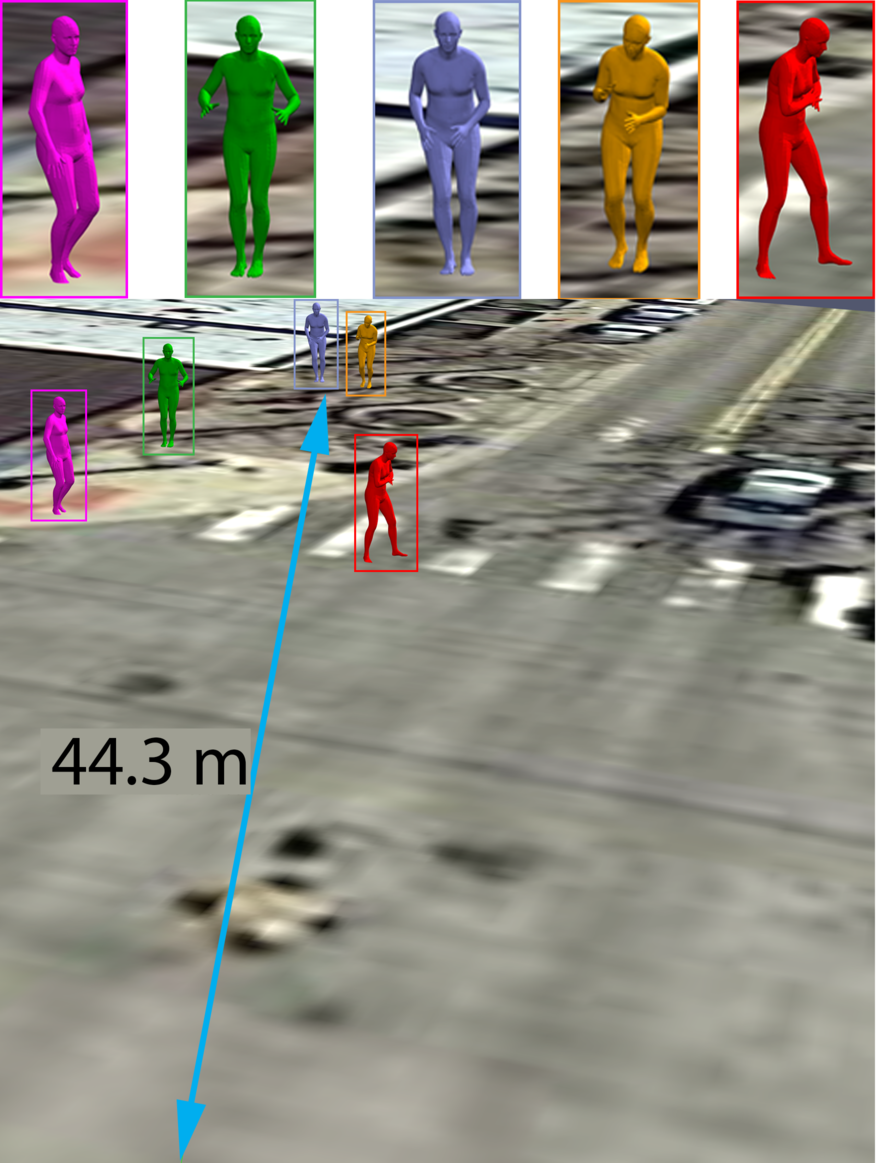}
    \caption{An example frame from the PedX dataset. \textit{Left}: A camera image with bounding boxes around the pedestrians within $5\text{-}45\mathtt{m}$ to the camera. \textit{Right}: A rendered image with the 3D human mesh models in metric scale.}
    \label{fig:intersection-example}
    \vskip -0.5cm
\end{figure}

Most AVs have cameras installed, so this data can serve as a primary source for human pose estimation using computer vision algorithms. In addition to cameras, LiDAR (Light Detection and Ranging) has become an essential component for AVs due to its precise depth measurements. This motivated the importance of capturing both modalities for this benchmark set of complex urban intersections.

Our dataset has the following unique properties. First, the data are gathered outdoors with real challenges such as varied lighting and weather conditions and the presence of occlusions. Second, the pedestrian data are collected at intersection length scales of up to $45$m range, which is relevant for deployment of pose estimation systems at application relevant scales. The captured scenes are also naturalistic. The pedestrians in our dataset are not actors, so they move and interact in a myriad of realistic ways. In addition, our dataset includes multi-person images capturing crowds of people. Lastly, we release multimodal pedestrian data including high resolution images and point clouds that are synchronously captured from stereo cameras and LiDAR sensors.

Annotations of our dataset also have distinctive features. All the annotated 3D pedestrians lie in a global metric-space coordinate frame, as opposed to many existing datasets that operate in hip joint or camera center relative coordinate frames. We stress the importance of determining where a person is in the 3D world so one can plan actions around them. In addition, our multimodal data frames are captured in minutes long sequences with unique tracking IDs for each pedestrian, enabling temporal reasoning.

The contributions of this paper are summarized as follows:
\begin{enumerate}[label=\arabic*.] 
    \item We release a publicly available large-scale multimodal pedestrian dataset with a rich set of 2D and 3D annotations. The dataset captures the real world challenges of urban intersections.
    \item We present an automatic method to obtain full 3D labels from 2D data, enabling labeling of in-the-wild images without mocap.
    \item Our automatic 3D labeling method is validated using a mocap system in a controlled outdoor environment that simulates pedestrians in urban intersections.
    \item Using state of the art algorithms for 2D annotations, our proposed approach allows generating 3D data in a completely unsupervised manner.
\end{enumerate}

We present this dataset to enable the study of 3D pose estimation while reasoning about pedestrian behavior around vehicles, particularly in crowded urban areas as depicted in Fig.~\ref{fig:intersection-example}. We see this as one of the first areas where human pose estimation can have a tremendous impact on safety and intelligence of mobile robot systems. Understanding the pose of road users affords information about activity, attention, and predictions of future position which are critical to safely navigate around humans.

\section{Related Work}
\label{sec:related-work}
\subsection{3D human pose estimation}
In many papers, 3D human pose estimation has been formulated as a problem of regressing 3D joint locations by directly extracting visual features from an image \cite{insafutdinov16ariv,popa2017deep,mehta2017vnect,zhou2017towards,zhou2016sparseness,mehtamonocular,chen20163d,tome2017lifting}, or by lifting 2D joint detector outputs to 3D joints in a camera relative frame~\cite{martinez_2017_3dbaseline,ramakrishna2012reconstructing,Bogo:ECCV:2016}.
To reduce the inherent ambiguity of 3D human pose estimation from a 2D image, the constraints on feasible human pose have additionally been considered~\cite{ramakrishna2012reconstructing,akhter2015pose}, or a deformable 3D human model such as Skinned Multi-Person Linear (SMPL)~\cite{SMPL:2015} has been used to be fit to known 2D joint locations~\cite{Bogo:ECCV:2016}.

Recently, pose representations other than sparse 3D joints have been used to formulate 3D human pose estimation. SMPL model parameters are directly estimated given dense 2D keypoints~\cite{lassner2017unite}, or  the parameters are predicted using end-to-end deep networks with the adversarial loss~\cite{kanazawa2018end}. UV parameterization of the 3D human body surface is estimated from the dense image-to-surface correspondence regression networks~\cite{Guler2018DensePose}. The per-pixel body depth map for each soccer player in a YouTube video is estimated using a neural network trained on synthetic data~\cite{rematas2018soccer}.

Most approaches take as input an image with a single person or a cropped image patch centered around a person, and return as output 3D pose in a root-relative coordinate frame where the camera is facing toward the person. While the joint estimation of 3D pose and virtual camera parameters have been proposed \cite{ramakrishna2012reconstructing, chen20163d}, the outputs are still not in real metric scale. Without knowledge of exactly how far away a person is, controlling a mobile robot safely will be challenging. Most approaches are also unable to handle multi-person images with a single pass of an algorithm. While DensePose~\cite{Guler2018DensePose} handles in-the-wild images with multiple people, dense pose outputs from a multi-person image cannot be converted into a common unified coordinate frame. 

A major impediment to predicting metric space pose for multiple people in a scene is the lack of a suitable dataset with reliable 3D annotations. Addressing this limitation is the focus of this paper.

\subsection{3D human pose datasets}
Large scale datasets have played an essential role in fueling recent progress in a variety of computer vision tasks. 
However, building a ground truth 3D human pose dataset in metric space is challenging since annotation in 3D is far more time consuming than the same task in 2D. 
To avoid manual labeling in 3D, mocap systems are typically used to obtain the ground truth 3D human pose~\cite{h36m_pami,sigal2010humaneva,ofli2013berkeley,cmu_mocap}.

The data captured with mocap has many limitations.
For instance, markers must be attached to subjects which makes images look unnatural. 
Moreover since mocap is typically restricted to constrained, mostly indoor areas, image backgrounds for mocap datasets have limited variability.
In addition the number of makers that can be tracked by mocap systems is limited, which restricts the number of subjects that can appear in any scene. 

To counter the limited variability in subject appearance, camera viewpoints, lighting conditions and image backgrounds, synthetic 3D datasets have been proposed~\cite{ghezelghieh2016learning,chen2016synthesizing}. While there have been attempts to improve the photo-realism of these synthetic data generation pipelines~\cite{varol2017learning,rogez2016mocap,mono-3dhp2017}, state-of-the-art synthetic images are still easy to distinguish from real images. Others have explored techniques to automatically do 3D labeling with limited human intervention ~\cite{Pavlakos:2017,lassner2017unite,huang2017towards}. 
For instance, some have taken advantage of a multiview camera setup to obtain reliable 3D annotations using optimization~\cite{Pavlakos:2017}.
Others explore fitting parameterized mesh models~\cite{Bogo:ECCV:2016} to monocular images or multiview images to collect the ground truth 3D labels~\cite{lassner2017unite,huang2017towards}
However, these methods provide 3D scale models for a virtual camera relative frame and typically for images from various 2D pose datasets cropped around a single detection.
This limits the potential utility of these methods while performing mobile robotic tasks safely.
In contrast, we construct 3D human pose models in metric space for crowded urban street intersections that include as many as $15$ pedestrians in a single image at distances as far as $45m$ from the camera.

\begin{figure*}[!t]
    \centering
    \includegraphics[width=1.0\linewidth]{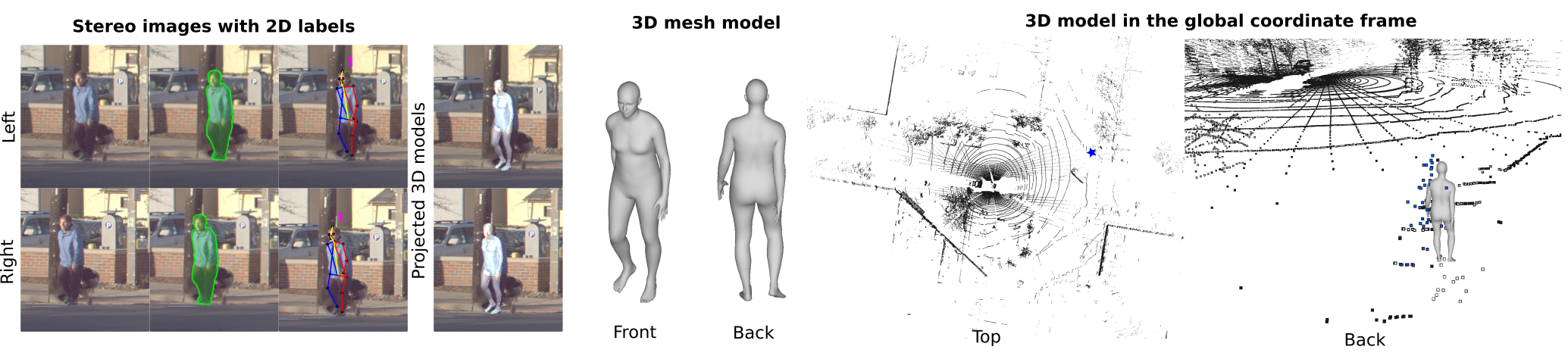}
    \caption{Visualization of our annotations. For each pedestrian, we provide 2D segmentation, 2D joint locations with visibility of 18 body joints, tracking ID, time-synced LiDAR points, and 3D mesh model localized into the global coordinate frame.}
\label{fig:annotation-examples}
\vskip -0.4cm
\end{figure*}
\begin{table}[ht!]
\centering
\caption{Statistics and characteristics of related datasets.}
\label{table:stat}
\setlength{\tabcolsep}{1.2pt}
\renewcommand{\arraystretch}{1.5}
\scalebox{0.77}{
\begin{tabular}{@{}ccccccccc@{}}
\toprule
 & \begin{tabular}[c]{@{}c@{}}num.\\images\end{tabular}
 & \begin{tabular}[c]{@{}c@{}}num.\\instances\end{tabular}
 & \begin{tabular}[c]{@{}c@{}}num.\\pixels\end{tabular}
 & \begin{tabular}[c]{@{}c@{}}3D\\points?\end{tabular}
 & video?
 & \begin{tabular}[c]{@{}c@{}}multi-\\person?\end{tabular}
 & real?
 & scene type\\ \midrule
H36M~\cite{h36m_pami} & 3,600,000 & 900,000 & $\approx$1M & Y & Y & N & Y & mocap indoor \\
HumanEva~\cite{sigal2010humaneva} & $\approx$80,000  & $\approx$80,000 &1-1.3M& Y & N & N & Y & mocap indoor \\
SURREAL~\cite{varol2017learning} & 653,6752 & 653,6752 & $\approx$80K & N & Y & N & N & indoor\\
UP-3D~\cite{lassner2017unite} & 8,515 & 8,515 & $\approx$300K &N & N & N & Y & indoor+outdoor\\
DensePose-COCO~\cite{Guler2018DensePose} & 33,929 & 131,129 & $\approx$300K & Y & N & Y & Y &indoor+outdoor\\
Ours & 10,152 & 14,091 & $\approx$10M & Y & Y & Y & Y & outdoor\\ \bottomrule
\end{tabular}
}
\vskip -.5cm
\end{table}
\newpage
\section{PedX Dataset}
\label{sec:dataset}
The PedX dataset contains more than $5,000$ pairs of high-resolution stereo images with $2,500$ frames of 3D LiDAR pointclouds. The cameras and LiDAR sensors are calibrated and time synchronized.
We selected three four-way stop intersections with heavy pedestrian-vehicle interaction. 
Cameras are installed on the roof of the car to obtain driver-perspective images. 
To cover all four crosswalks at an intersection, the images were captured by two pairs of stereo cameras -- one pair facing forward and another facing the incoming road from the left. 
Our dataset includes more than $14,000$ pedestrian models with a distance of 5-45\texttt{m} from the cameras and we provide reliable 2D and 3D labels for each instance. Table~\ref{table:stat} presents statistics of our dataset in comparison to other publicly available 3D human pose datasets.

Instance-level 2D segmentations and body joint locations are manually labeled for all images by instructed annotators. 
We also provide the unique tracking ID for each instance across the frames if the pedestrian appears in consecutive frames. 
A 2D segmentation is labeled by outlining a single connected polygon to cover the entire visible area of an object. 
To label the keypoints, $18$ body joints are selected including $4$ facial components. 
In cases where some keypoints are invisible due to occlusion by other objects in the scene or self-occlusion, annotators made a reasonable guess about the joint location, but also indicated the degree of occlusion.

We use the SMPL~\cite{SMPL:2015} parameterization to represent our 3D annotations which consist of shape and pose parameters and additionally estimate the global location and orientation of the instance in the global coordinate frame. 
The best fit 3D SMPL models were computed using the 2D annotations from a pair of stereo images and LiDAR points. 
The obtained 3D models encode pose, shape, and global location in 3D metric space without scale ambiguities.
Fig.~\ref{fig:annotation-examples} illustrates 2D and 3D annotations from our dataset.
The automatic algorithm to determine the optimal SMPL parameters is discussed in the next section.

\section{Multi-modal 3D Model Fitting}
\label{sec:optimization}
We perform 3D model fitting on pedestrians in a stereo-LiDAR sequence. In contrast to the previous work~\cite{Bogo:ECCV:2016} that fits a 3D model to a single frame at a time, our approach optimizes over a sequence of stereo images and LiDAR points. We begin by per-instance model fitting which is then extended to optimize over a sequence of instances using an iterative method. 
We also propose multi-modal and temporal priors.
Note that we use a gender-neutral model for model fitting. Before initiating the model fitting pipeline, we preprocess the LiDAR data to identify regions containing potential pedestrians in 3D space. Using 2D segmentation labels for stereo images with known transformation between LiDAR and camera coordinate frames, we perform the point cloud labeling of each pedestrian instance. 

\subsection{Fitting to a single instance (at a single time step)}
We begin by performing 3D model fitting to a single instance at a single time step. 
For each pedestrian instance, we are given 2D joint locations $\bm{x}_l$ and $\bm{x}_r$, and 2D segmentations $S_l$ and $S_r$ for each stereo image. 
We also have sparse 3D points corresponding to the instance. 
To find the pose $\bm\theta$, shape $\bm\beta$, and 3D global position $\bm{t}$ that best fit to the instance, we formulate the problem as:
\begin{align}
    \underset{\bm\theta,\bm\beta,\bm{t}}{\minimize}\:
        E_I\left(\bm\theta,\bm\beta,\bm{t}\right)\label{eq:per-frame}
\end{align}
where $E_I = E_J + E_{3d} + E_P + E_{T} + E_{D}$ represents the sum of multiple energy terms.
We verify the effectiveness of each energy term through ablative experiments described in Sec.~\ref{sec:ablation}. $E_J$ is the sum of robust 2D reprojection error~\cite{Bogo:ECCV:2016} for both left and right images, $E_P$ is the prior term, $E_{3d}$ is the 3D Euclidean distance term between visible SMPL vertices and the LiDAR points, $E_T$ is the translation term to constrain the 3D model location, and $E_D$ is the heading direction term to constrain the body orientation:
\begin{align}
    & E_T(t) = \left\|\bm{t}-\bm{t}_0\right\|_2^2,\label{eq:ET}\\
    & E_D(\bm\theta) = \|f(\bm\theta) - \bm{d}\|_2^2,\label{eq:ED}\\
    & E_{3d} = 
        \textstyle \frac{1}{N_v} \sum_{i}^{N_v}
        \underset{j}{\min} \|\bm{X}_i - \bm{V}_j\|_2^2,\label{eq:E3D}
\end{align}
where $\bm{t}_0$ is the mean of 3D points, $f$ is a function to convert the axis-angle representation of body orientation to xyz- directional vector, $\bm{d}$ is a known heading direction vector, $\bm{X}_i$ is the i-th LiDAR point, $N_v$ is the total number of 3D points that belong to the instance, and $V_j$ is the j-th point of SMPL model vertices.

\subsection{Fitting to a sequence of single instances}
To fit 3D models to a sequence of detections, we develop \textit{shape} and \textit{temporal consistency} constraints across frames in addition to the per-frame constraints.

\subsubsection{Global shape consistency}
Suppose one pedestrian appears in $N$ consecutive frames with full 2D labels.
In this instance, while pose parameters and translations change, the shape parameters should remain unchanged across the sequence. 
To find the pose, shape parameters and translations, we formulate the problem as:
\begin{align}
    \underset{\bm\theta,\bm\beta,\bm{T}}{\minimize}\:&
    E_{seq}\left(\bm\Theta,\bm\beta,\bm{T}\right)
\end{align}
where $\bm\Theta = \{\bm\theta_1,\ldots,\bm\theta_N\}$ and $\bm{T} = \{\bm{t}_1,\ldots,\bm{t}_N\}$ are the set of pose parameters and the set of translations for all $N$ frames. $\bm\beta$ is the shape parameters shared by all frames where $\bm\beta = \bm\beta_1 = \cdots = \bm\beta_N$. 
Optimizing over the entire sequence is challenging due to the high dimension of the decision variables. 
Since the objective $E_{seq}$ is separable in terms of a per-frame objective, we decompose this large problem into a set of smaller problems. 
We rewrite the unconstrained minimization problem over the entire sequence by introducing the consensus variable $\bm\beta$:
\begin{gather}
    \underset{
        \bm\theta_{1:N},\bm\beta_{1:N},
        \bm{t}_{1:N},\bm\beta}{\minimize}\:
    \textstyle{\sum_{k=1}^{N}}
    E_{I,k}\left(\bm\theta_k,\bm\beta_k,\bm{t}_k\right)
    \notag\\
    \subjectto\:\:\bm\beta_k - \bm\beta = 0,
    \:k\in\{1,\ldots,N\}
\end{gather}
where $k$ denotes the frame ranging from $1$ to $N$ frames in a sequence. This optimization is a constrained minimization problem with a separable objective function and multiple constraints that require that each per-frame shape parameter $\bm\beta_k$ is equal. 
The advantage of introducing a consensus variable $\bm\beta$ is that we can enforce all the frames to have common shape parameters while exploiting parallelism. 
We solve the problem by using the alternating direction method of multipliers (ADMM)~\cite{boyd2011distributed}. 
The augmented Lagrangian to be minimized is:
\begin{align}
    \small
    \mathcal{L}_{\rho}
    \left(\bm\beta, \bm\beta_k; \bm\Theta, \bm{T}\right)
    =&
    {\textstyle \sum_{k=1}^{N}}
    E_{I,k}\left(\bm\beta_k;\bm\Theta,\bm{T}\right)\notag\\
    &+ 
    \bm{y}_k^T (\bm\beta_k - \bm\beta)
    + \frac{\rho}{2}\left\|\bm\beta_k-\bm\beta\right\|_2^2
\end{align}
$\bm{y}_k$ is the dual variable for $\bm\beta_k$ and $\rho$ is a positive constant that is experimentally selected. $\rho$=2 was used for the results reported in this paper. The objective $\mathcal{L}_{\rho}$ is optimized using an alternating optimization for the local and global shape parameters with variables $\{\bm{u}_k\}_{k=1}^N$ where $\bm{u}_k=\textstyle\frac{\bm{y}_k}{\rho}$. 
The update equations at each iteration are as follows:
\begin{align}
    \bm\beta_k^{t+1}
    &:=
    {\textstyle\argmin_{\bm\beta_k}}\:
    \scriptstyle{E_{I,k}\left(\bm\beta_k;\bm\Theta,\bm{T}\right)}
    + \textstyle{\frac{\rho}{2}}\left\|
    \bm\beta_k - \bm\beta^t + \bm{u}_k^t\right\|_2^2 \label{eq:primal-update}\\
    \bm\beta^{t+1}
    &:=
    \textstyle{\frac{1}{N}\sum_{k=1}^N}
    \left(\bm\beta_k^{t+1} + \bm{u}_k^{t}\right)\\
    \bm{u}_k^{t+1}
    &:=
    \bm{u}_k^{t} + \bm\beta_k^{t+1} - \bm\beta^{t+1}
\end{align}
\noindent
We perform the synchronous update for the global shape parameters. 
The iteration is stopped when $\|\bm\beta_k^{t+1}-\bm\beta^{t+1}\|_2 < 0.05$ and $\rho\|\bm\beta^t - \bm\beta^{t+1}\|_2 < 0.05$ or the maximum iteration is reached. 
\eqref{eq:primal-update} is similar to per-frame minimization in \eqref{eq:per-frame} with an additional term in the objective function.

\subsubsection{Temporal pose prior}
\label{sec:temporal-prior}
In addition to enforcing per-frame shape parameters to share the common values across the sequential frames, we propose a temporal pose prior to give a penalty to unlikely sequences of poses.
A 72-dimensional pose vector consists of xyz rotation angles of 23 joints relative to each of their parent nodes, plus the orientation of the root hip in 3D angle-axis representation. 
We observe that the difference between the pose vectors from two consecutive frames is small and has some patterns common to individual joints. 
Especially for pedestrians at the intersection, many of them are involved in walking or other actions at slow speed. 
Since the difference in rotation angles can also be affected by the translation velocity of pedestrians, we define a 75-dimensional vector with the first 3 components consisting of the difference in body translation: $\Delta\bm{x} = (\Delta\bm{t}, \Delta\bm\theta)$.

We fit a Gaussian mixture models with $10$ distinct distributions for the pose difference vectors using the CMU mocap dataset~\cite{cmu_mocap}. 
Since the frame rate of the mocap and our data capture is different, we use the mocap frame rate when estimating this pose difference vector. 
We include the negative log of this multivariate normal probability distribution as part of the prior in the objective term:
\begin{gather}
    \Delta\bm{x}^t = \bm{x}^{t} - \bm{x}^{t-1}
    \:\sim\:
    \textstyle\sum_{i}^N w_i\:
    \mathcal{N} \left(
    \Delta\bm{x}^t; \mu_i, \Sigma_i
    \right)\\
    \resizebox{.88\hsize}{!}{
        $E_{tp}(\bm{t}_k,\bm\theta_k;\bm{t}_{k\text{-}1},\bm\theta_{k\text{-}1})
        = -\log\textstyle
        \sum_{i}^N w_i
        \mathcal{N} \left(
        \Delta\bm{x}^t; \mu_i, \Sigma_i
        \right)\label{eq:Etp}$
    }
\end{gather}
where $\mu_i$, $\Sigma_i$ are the mean and covariance of the pose difference vector $\Delta x$ and $w_i$ is the weight for the i-th Gaussian mixture component.

\begin{figure}[t]
\centering
\subfloat[]{
    \includegraphics[height=0.23\columnwidth]{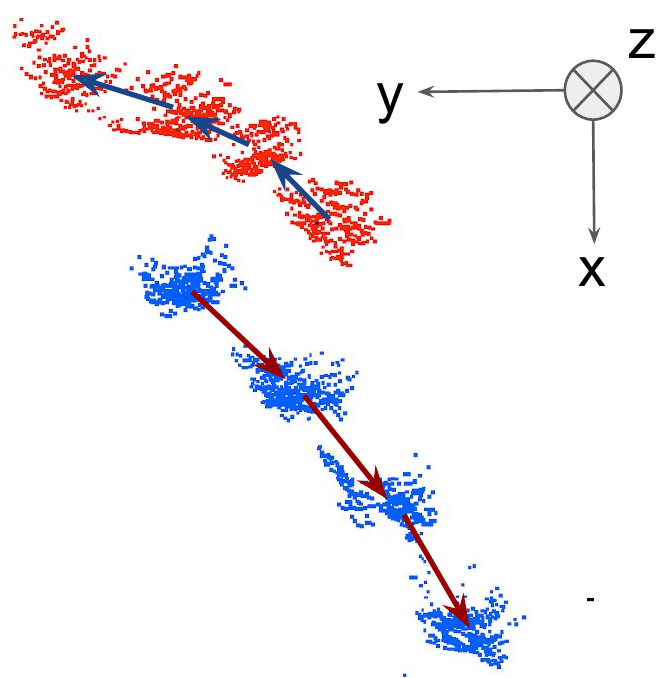}
    \label{fig:pcd-traj}
}\hskip -10pt
\subfloat[]{
    \includegraphics[height=0.24\columnwidth]{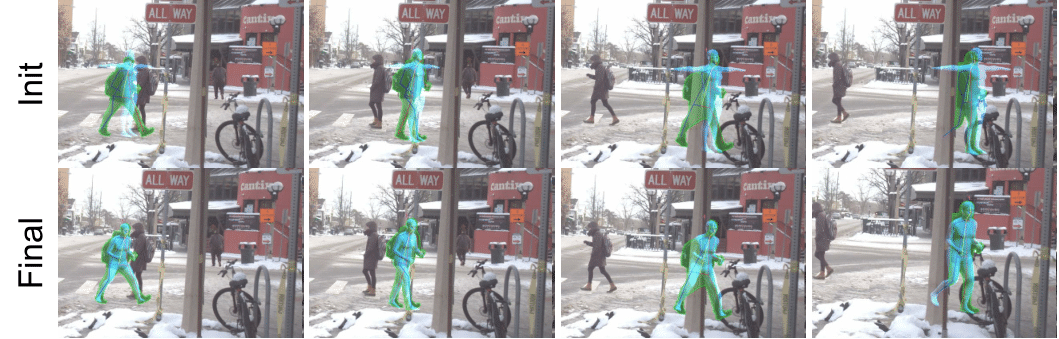}
    \label{fig:init-orient}
}
\vskip -0.2cm
\caption{(a) Overhead view of the pointcloud trajectories of pedestrians in the global frame. (b) Initialized body orientation based on heading direction (top) and the final 3D models after the iterations (bottom).}
\vskip -0.4cm
\end{figure}

\subsection{Initialization}
One of the challenges while fitting a 3D model is estimating a body orientation with the incorrect sign~\cite{huang2017towards}. To avoid getting a flipped model, we compute the heading direction of each pedestrian from a sequence of LiDAR points and use it to initialize the body orientation via a 3-dimensional angle-axis representation. 
We assume that pedestrians never move backwards (which held true in our data capture).

Fig.~\ref{fig:pcd-traj} shows some example trajectories from LiDAR points, and Fig.~\ref{fig:init-orient} shows projected 3D model onto the left images at four sequential time frames after the initialization.
When a sequence is only a single frame, we find the initial body orientation with the template pose by minimizing the stereo reprojection error $E_J$ only using torso body joints and the translation error $E_T$ around the mean LiDAR points.

\section{Experiments}
\label{sec:experiment}
The data in our dataset is captured at complex outdoor urban intersections where pedestrian-to-camera distances are large (5-45m), with multiple subjects who are often heavily occluded.
In contrast, publicly available 3D datasets rely on mocap systems \cite{ofli2013berkeley,h36m_pami,sigal2010humaneva} guarantee less than a few $\mathtt{mm}$ accuracy for a \emph{single} subject appearing within a controlled \emph{indoor} capture volume of a \emph{few meters} in radius. 
Given these distinctions, the accuracy of our dataset needs to be evaluated under these realistic conditions. 
While comparing the accuracy of our proposed approach with a mocap system would reliably validate our dataset, mocap systems cannot be practically setup at urban intersections.

We address this challenge by leveraging the fact that our proposed approach only requires 2D labels and LiDAR data without considering image features. 
The key factors affecting our approach include the density of the LiDAR returns due to the large distance from the capture vehicle, occlusions by other vehicles and other pedestrians which affect the LiDAR segmentation, the precision of the manual annotation when the pedestrian occupies a small portion of the image and the calibration between LiDAR and camera. 
The lighting conditions, background appearance, clothing and weather conditions do not affect our approach as we operate on manually labeled joint locations. 
Therefore, we collect and annotate an evaluation dataset in a controlled outdoor environment with a mocap system and get manual 2D labels while replicating the vehicle to target distances and the clutter and occlusion of the intersection data. 
By showing that the 3D labels generated by our method, using the same hand annotation process, is comparable to the mocap ground truth, we independently verify the fitting approach and the hand annotation process against a traditional mocap source.

\subsection{Data Verification}
\label{sec:data-verification}
\subsubsection{Evaluation dataset}
We use the PhaseSpace mocap system with active LED markers which can be used in outdoor environments. The subject wears a suit with markers placed around body parts and repeats actions such as walking, jogging, and waving that are common for pedestrians. 
The capture vehicle was parked about $20\mathtt{m}$ away from the mocap setup. To replicate typical occlusions we parked another car between the capture vehicle and the mocap setup as well as having groups of pedestrians walking. We selected $626$ frames and obtained manual 2D labels for the images.
Since the visual appearance does not affect the evaluation, we restrict the evaluation to a single subject with a single background and focus more on variation in poses and occlusions.

\subsubsection{Evaluation metric}
The 3D mean per joint position error (MPJPE) is a standard metric to evaluate pose estimation algorithms which is a mean over all joints of the euclidean distance between ground truth and prediction. In cases where the prediction is not in metric space, the error is computed for a root-relative coordinate frame after allowing a similarity transform to register the prediction to the ground truth. In cases where the prediction is in metric space, we compute MPJPE in global coordinate frame without any registration. We further report the per joint position errors for both frames. Note that, given the geometry of capture, markers can get completely occluded from the mocap system. Consequently, not all joints are visible in all frames. Moreover, some methods may not predict invisible joints. Therefore, we take the weighted mean while computing the MPJPE where the weight is equal to the number of frames in which the joint was visible in the ground truth and was predicted. %

\subsubsection{Baseline Methods}
We consider three different families of baseline methods. 
First, we consider a method that predicts 3D joint coordinates (up to scale) directly from 2D images~\cite{zhou2017towards}. Second, we consider methods that take manual 2D joint annotations as inputs ~\cite{martinez_2017_3dbaseline,Bogo:ECCV:2016}. 
We evaluate these methods in the root relative frame alone.
Third, we consider three naive baselines that use stereo geometry information. 
As we have 2D joint locations for a calibrated rectified stereo pair of images, we directly triangulate these 2D joint locations for visible joints.
We refer to this method as \emph{Triangulation}. For the second naive baseline, we use disparity values and 2D joint locations in the left image for the visible joints and the previous triangulation result for invisible joints. 
We refer to this as \emph{Left+disp}. 
Finally, we consider a baseline that modifies an existing technique \cite{Bogo:ECCV:2016}, this approach uses the calibrated camera parameters and the estimated skeletons which are scaled to metric space by using the average disparity values at the visible joint locations. 
We refer to this as \emph{SMPLify~\cite{Bogo:ECCV:2016}+disp}.

\begin{table}[t]
\raggedright
\caption{3D MPJPE in root-relative coordinate frames.}
\vskip -0.2cm
\label{table:mpjpe-local}
\scalebox{0.78}{
\setlength{\tabcolsep}{2.2pt}
\renewcommand{\arraystretch}{1.2}
\begin{tabular}{@{}lccccccccccccc|c@{}}
\bottomrule
\begin{tabular}[c]{@{}c@{}}Relative\\($\mathtt{mm}$)\end{tabular}
& rknee & lknee & rankl & lankl & rsho & lsho & relb & lelb & rwri & lwri & head & neck & hip & mean \\
\bottomrule
\cite{zhou2017towards}
& 113 & 153 & 203 & 184 & 141 & 120 & 130 & 132 & 203 & 194 & 134 & 96 & 88 & 147 \\
\midrule
\cite{martinez_2017_3dbaseline} 
& 107 & 116 & 172 & 159 & 159 & 160 & 142 & 127 & 197 & 178 & 84 & 88 & 87 & 137 \\ 
\cite{Bogo:ECCV:2016}$\quad\quad\quad$ 
& \textbf{103} & 89 & 139 & 158 & 77 & 59 & \textbf{74} & 71 & 144 & 145 & 85 & 35 & \textbf{87} & 97 \\
\midrule
Ours
& 104 & \textbf{71} & \textbf{126} & \textbf{136} & \textbf{62} & \textbf{50} & 83 & \textbf{67} & \textbf{118} & \textbf{130} & \textbf{66} & \textbf{29} & 106 & \textbf{88} \\ 
\bottomrule
\end{tabular}
}
\\\vskip 0.3cm
\raggedleft
\caption{3D MPJPE in global coordinate frames.$\quad\quad$}
\vskip -0.2cm
\label{table:mpjpe-global-full}
\scalebox{0.78}{
\setlength{\tabcolsep}{1.2pt}
\renewcommand{\arraystretch}{1.2}
\begin{tabular}{@{}lccccccccccccc|c@{}}
\bottomrule
\begin{tabular}[c]{@{}c@{}}Global\\($\mathtt{mm}$)\end{tabular}
& rknee & lknee & rankl & lankl & rsho & lsho & relb & lelb & rwri & lwri & head & neck & hip & mean \\
\bottomrule
Triangulation
& 1178 & 1307 & 1617 & 1489 & 1205 & 1164 & 1130 & 1168 & 1206 & 1111 & 842 & 1029 & 1111 & 1194 \\
Left+disp
& 766 & 977 & 1229 & 1018 & 702 & 648 & 756 & 766 & 825 & 789 & 482 & 437 & 972 & 794 \\
\cite{Bogo:ECCV:2016}+disp
& 535 & 576 & 627 & 591 & 620 & 574 & 561 & 570 & 652 & 594 & 588 & 587 & 598 & 593 \\
\midrule
Ours
& \textbf{205} & \textbf{179} & \textbf{250} & \textbf{255} & \textbf{183} & \textbf{177} & \textbf{187} & \textbf{182} & \textbf{221} & \textbf{204} & \textbf{169} & \textbf{155} & \textbf{161} & \textbf{194} \\ 
\bottomrule
\end{tabular}
}
\vskip -0.5cm
\end{table}

\subsubsection{Accuracy} 
Table~\ref{table:mpjpe-local} and Table~\ref{table:mpjpe-global-full} summarize the results for root-relative and global frames respectively. Although fair comparisons can only be done between methods separated by horizontal lines, the objective of these tables is to highlight that the current state-of-the-art still has room for improvement and the utility of the proposed dataset in closing this gap. 
The proposed approach achieves lower MPJPE for majority of joints in the root-relative frame. This is expected as our approach leverages additional information including LiDAR returns, temporal priors as well as stereo annotations. 
The gains while using this additional data are most prominent in the global coordinate frame as no registration is involved and consequently global translation, orientation and scale errors can be seen in the global MPJPE. While naive baselines such as triangulation and left+disp perform poorly as they do not leverage any prior about proportions of a typical human skeleton, SMPLify+disp which leverages priors about human skeletons still suffers from large errors. In contrast, our proposed approach achieves an MPJPE of $194\mathtt{mm}$ for an average camera-to-pedestrian distance of $20\times 10^3\mathtt{mm}$.

\subsection{Ablation Study}
\label{sec:ablation}
Table~\ref{table:ablation} summarizes the results for using different subsets of energy terms in the optimization.
    Note that $E_{J,l}$ and $E_{J,r}$ represents the reprojection error on left and right images. $E_T$, $E_{3D}$ and $E_{tp}$ are defined in Sec.~\ref{sec:optimization}. 
Each column shows per-joint errors in root-relative frame except the last column which shows the MPJPE in global frame.

\subsubsection{Effect of stereo}
To see how using stereo reprojection error affects the resulting 3D models, we compute reprojection errors for only the left images (i.e. row 4 of Table~\ref{table:ablation}) and for both stereo images (i.e. row 5 of Table~\ref{table:ablation}).  
Stereo imagery reduced both global translation error and root-relative pose error as it reduces the depth ambiguities that exist for the monocular approach. 
The second row of Fig.~\ref{fig:ablation-local} shows the results from monocular approach which estimates 3D models from the left images. 
Notice that in several frames, the legs are swapped or the body orientation is estimated incorrectly.
Fig.~\ref{fig:ablation-occlusion} presents additional results with occlusions, and illustrates similar limitations of the existing approach. 
We can see that the projection of the estimated 3D models onto the right image does not align exactly.
Estimating 3D pose from 2D joints is an inherently ill-posed problem because there exist many feasible body configurations. 
Using stereo information provides more constraints during the optimization and overall, reduces such ambiguities. 
Furthermore, when some joints are occluded in a single image, when using stereo pairs the second image of the pair may observe these joints which can reduce the uncertainty and produce better 3D models.

\subsubsection{Effect of using LiDAR points}
Although stereo images provide reasonable depth estimation in a global coordinate frame, the translation error may be too large since an error of a few pixels during labeling may create a large resulting error in fit.
To place 3D models at the correct location in metric space, we include LiDAR information as a form of the translation prior $E_T$. 
The translation prior term localizes the 3D models at the distance observed by the LiDAR. 
As shown in the first two rows in Table~\ref{table:ablation}, adding this translation prior provides an improvement in estimating global 3D pose.

The translation prior only constrains the location of the root joint (hip) in the 3D metric space. 
Therefore, depth ambiguities in other parts of the body may still exist, especially when the subject appears sideways with respect to the camera.
Adding the 3D distance term $E_{3D}$ helps to adjust the pose or body orientation to fit to the observed LiDAR points. 
Consequently the mean column in the root-relative frame significantly reduces from row 2 to row 4.

\subsubsection{Temporal prior}
The temporal prior term, $E_{tp}$, penalizes unlikely transitions of poses and translations between consecutive frames.
It also makes the resulting 3D pose between consecutive frames appear smooth. 
In Table~\ref{table:ablation}, row 2 and row 3, row 4 and row 5 show the errors without and with the temporal prior, respectively. 
By adding this term we obtain similar values for the root-relative errors, and achieve lower global error.
Another advantage of the temporal prior is that it makes the model robust to 2D labeling noise for the occluded joints. 
Fig.~\ref{fig:ablation-occlusion} illustrates examples with severe occlusions. 
It is likely that the 2D body joint labels are inconsistent across the frames under severe occlusion, and that affects the estimated pose. 
As seen in rows 4 and 5 of Fig.~\ref{fig:ablation-occlusion}, the resulting 3D poses are smoother when this temporal prior term is included. 
However, if the weight of temporal prior is set too high, the transition of poses between the frames can become too restricted.

\subsubsection{Global shape consistency}
\begin{figure*}[t]
    \centering
    \includegraphics[width=1.0\textwidth]{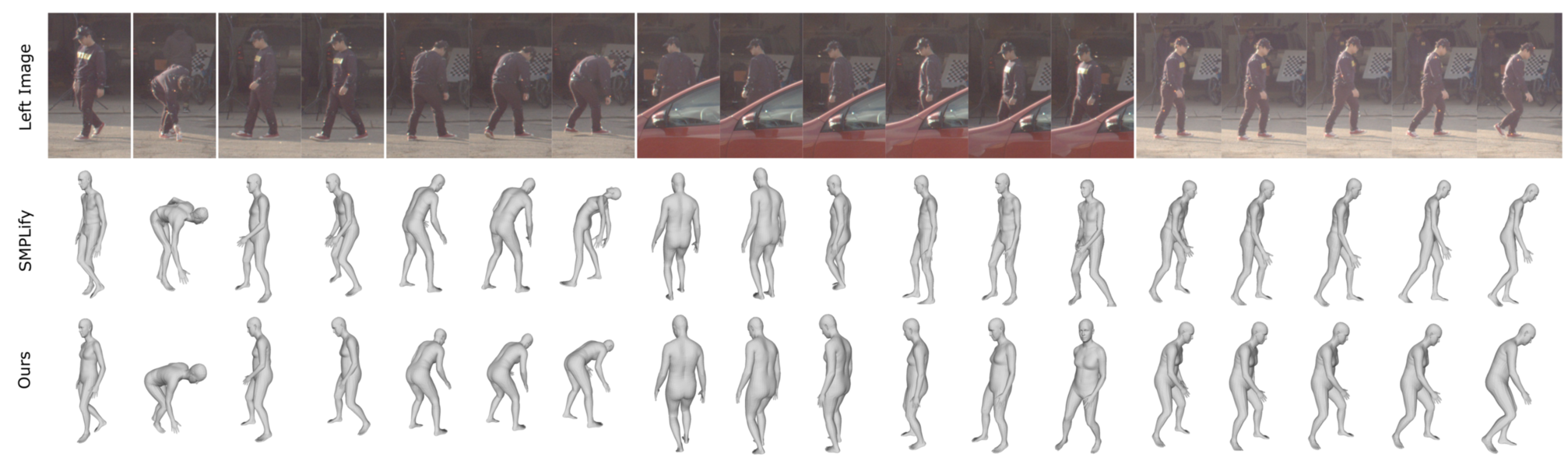}
    \vskip -.4cm
    \caption{Results from monocular SMPLify and from our method.}
    \label{fig:ablation-local}
    \includegraphics[width=1.0\textwidth]{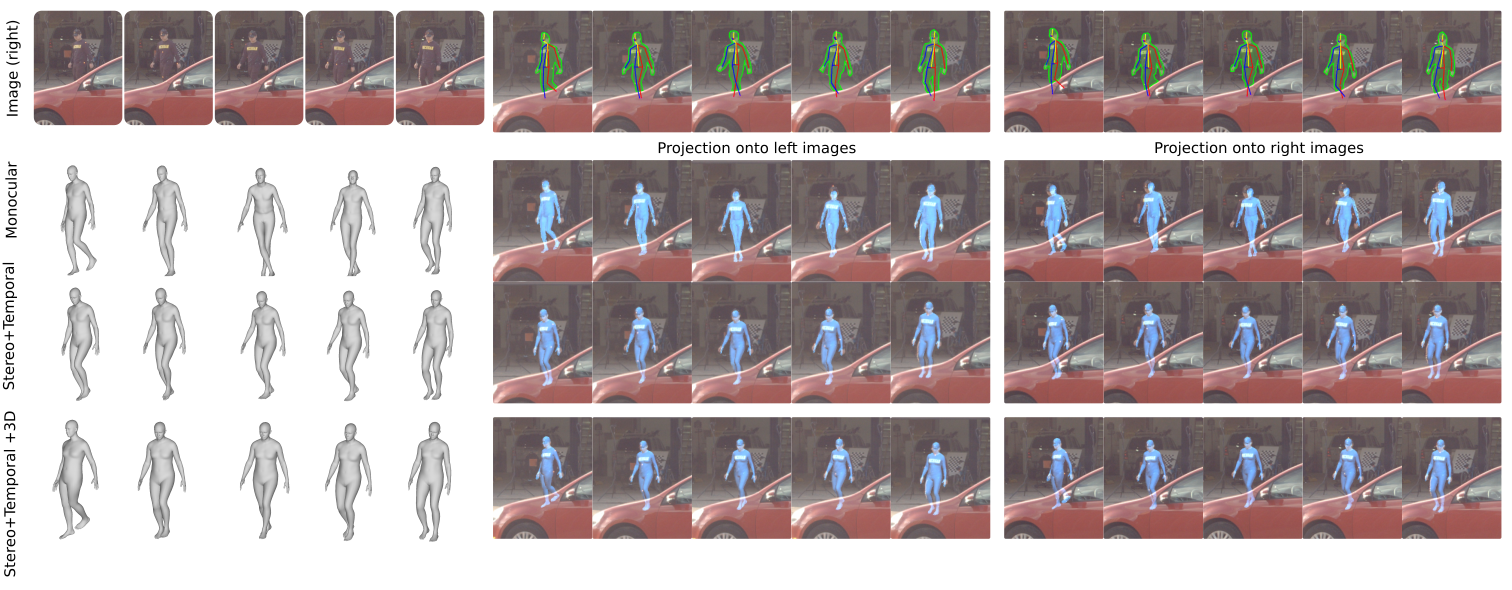}
    \vskip -.3cm
    \caption{Results from using different cost terms.
    The manual annotation for both stereo images is shown and the resulting models in the temporal walking sequence with and without the proposed novel terms. 
    Note the inclusion of the additional cost terms helps to disambiguate complicated and occluded poses where the original fitting approach struggles to deal with the depth ambiguity particularly near the lower limbs that are heavily occluded in this example.}
    \label{fig:ablation-occlusion}
    \vskip .2cm
    \includegraphics[width=1.0\textwidth]{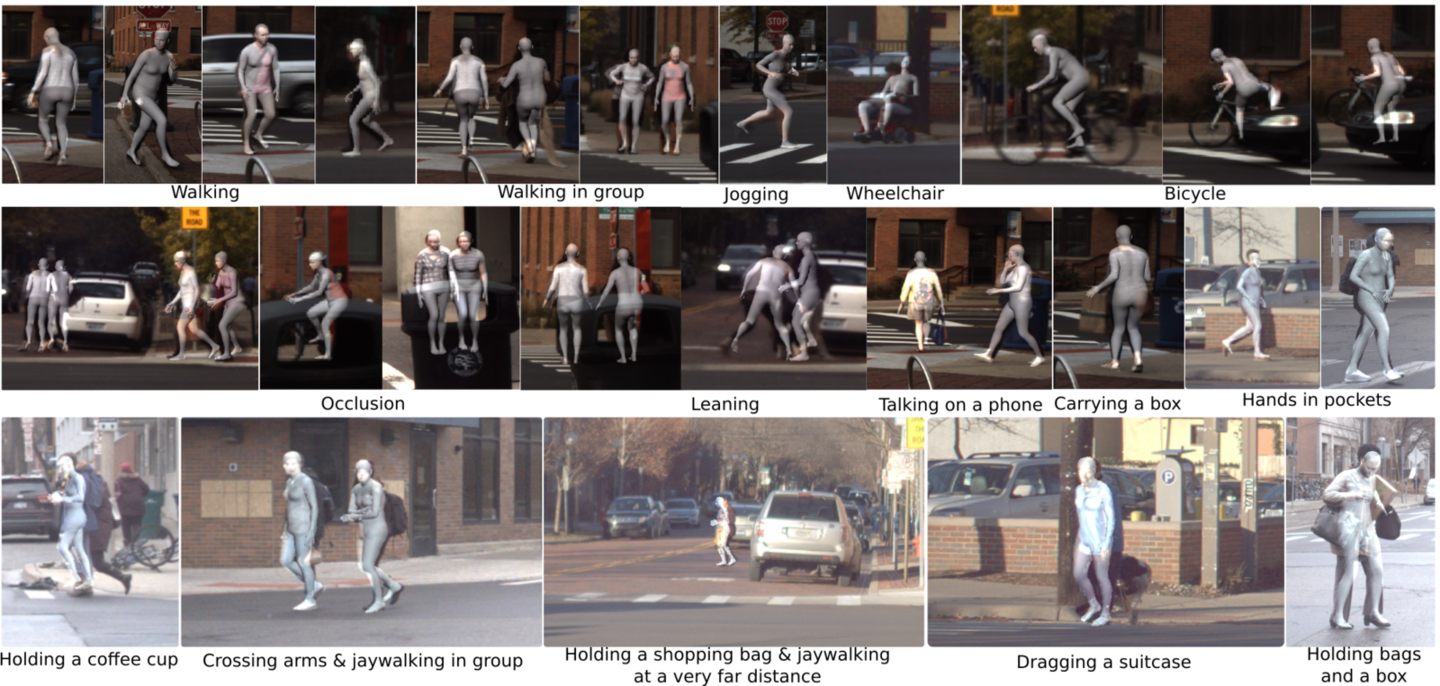}
    \caption{Representative samples from our dataset illustrates the utility of the dataset. The 3D models from our automatic labeling method are rendered onto the image to show the accuracy of the labels in our dataset even for challenging conditions.}
    \label{fig:pedx1-examples}  
    \vskip -.5cm
\end{figure*}

Fig.~\ref{fig:ablation-local} compares the results from SMPLify~\cite{Bogo:ECCV:2016} with those from our method which enforces consistency of the global shape parameters across multiple frames.
As expected using the global shape consistency constraint produces more consistent shapes across the frames, while the resulting 3D models from SMPLify, which only uses per-frame information, looks inconsistent. 
Moreover, the models are too skinny especially when the desired pose is far from the template pose.

\subsection{Effect of Noisy 2D labels}
\label{sec:autolabel}
The size of manually labeled datasets are always limited by the time consuming and costly annotation process.
This typically restricts the number of frames that can be annotated to only a few thousand.
Luckily, recent state-of-the-art methods have made great progress in 2D tasks such as segmentation, 2D joint detection and tracking. In this section, we use the segmentations from a pre-trained Mask R-CNN~\cite{he2017mask} and 2D joint detections from OpenPose~\cite{cao2017realtime} in place of manual labels on our evaluation dataset. Since there is only one subject wearing the mocap suit, the tracking ID is trivially obtained. Table~\ref{table:autolabel} shows the 3D MPJPE errors for using the 2D estimates and using ground truth labels. Although the accuracy decreases, the method still achieves greater accuracy than ad-hoc versions of monocular methods shown in Table~\ref{table:mpjpe-global-full}. This illustrates that our proposed approach is robust to noisy 2D labels and could potentially be scaled for larger datasets using state-of-the-art multi-object tracking algorithms along with segmentation and 2D joint detection networks in urban scenes.

\begin{table}[t]
\centering
\caption{Ablation study on energy terms. Each column shows per-joint errors in \texttt{mm} in root-relative frames. The last column shows the MPJPE in global frame.}
\vskip -0.2cm
\label{table:ablation}
\setlength{\tabcolsep}{1.2pt}
\renewcommand{\arraystretch}{1.1}
\scalebox{0.74}{
\begin{tabular}{@{}c@{\hskip -0.5pt}c@{\hskip -0.5pt}c@{\hskip -0.5pt}c@{\hskip -0.5pt}c@{\hskip -0.5pt}c@{\hskip -0.5pt}cccccccccccccc|c@{}}
\bottomrule
&$E_{J,l}$ &$E_{J,r}$ & $E_T$ & $E_{3D}$ & $E_{tp}$
& rknee & lknee & rankl & lankl & rsho & lsho & relb 
& lelb & rwri & lwri & head & neck & hip & mean & global\\ \hline
1&\checkmark&\checkmark & && 
& 100 & 79 & 128 & 145 & 66 & 54 & 85 & 66 & 147 & 165 & 72 & 38 & 134 & 98 &1050\\ 
2&\checkmark&\checkmark & \checkmark && 
& 98 & 76 & \textbf{124} & 143 & 68 & 55 & 84 & 66 & 141 & 157 & 73 & 37 & 121 & 95& 277\\
3&\checkmark&\checkmark & \checkmark & & \checkmark
& 117 & 76 & 126 & 138 & 64 & 50 & 81 & \textbf{64} & 123 & 137 & 68 & 34 & 116 & 91 & 248\\
4&\checkmark&\checkmark & \checkmark & \checkmark &  
& 104 & 72 & 125 & \textbf{135} & \textbf{62} & \textbf{49} & 83 & 67 & 122 & 134 & \textbf{65} & \textbf{29} & 107 & \textbf{88} & 250\\
5&\checkmark&& \checkmark & \checkmark & \checkmark 
& \textbf{89} & 84 & 132 & 146 & 68 & 57 & \textbf{74} & 65 & 143 & 158 & 80 & 38 & 122 & 97 & 240\\
6&\checkmark&\checkmark & \checkmark & \checkmark & \checkmark 
& 104 & \textbf{71} & 126 & 136 & 63 & 50 & 83 & 67 & \textbf{118} & \textbf{130} & 66 & 30 & \textbf{106} & \textbf{88} &\textbf{194} \\
\bottomrule
\end{tabular}}
\vskip 0.2cm
\caption{Comparison of 3D MPJPE in root-relative coordinate frame for automatic and manual 2D labels. The last column is the MPJPE in global coordinate frame (in $\mathtt{mm}$).}
\vskip -0.2cm
\label{table:autolabel}
\scalebox{0.80}{
\setlength{\tabcolsep}{1.2pt}
\renewcommand{\arraystretch}{1.1}
\hskip -0.2cm
\begin{tabular}{@{}lcccccccccccccc|c}
\bottomrule
2D labels
& rknee & lknee & rankl & lankl & rsho & lsho & relb & lelb & rwri & lwri & head & neck & hip & mean & global \\
\hline
\cite{cao2017realtime}+\cite{he2017mask}
& \textbf{104} & 89 & 143 & 156 & 81 & 61 & 89 & 75 & 143 & 134 & 112 & 45 & \textbf{59} & 99 & 224 \\ 
GT 2D joints
& \textbf{104} & \textbf{71} & \textbf{126} & \textbf{136} & \textbf{62} & \textbf{50} & \textbf{83} & \textbf{67} & \textbf{118} & \textbf{130} & \textbf{66} & \textbf{29} & 106 & \textbf{88} & \textbf{194}\\ 
\bottomrule
\end{tabular}
}
\vskip -0.5cm
\end{table}

\subsection{Qualitative Results}
\label{sec:qualitative-results}
Fig.~\ref{fig:pedx1-examples} shows some representative examples from \textit{PedX} that illustrate the uniqueness and variety of our dataset. Our dataset covers various actions and poses that are frequently encountered at intersections. Examples include walking, jogging, waving, using a phone, cycling, carrying objects, and talking. Occlusions are another challenge in estimating 3D pose. Our dataset contains many pedestrian instances with severe occlusions by surrounding objects or by other pedestrians. In addition, most frames of the dataset contain more than one pedestrian at a time. Our dataset also contains different weather conditions and rare occurrences such as people in wheelchairs or pedestrians jaywalking.

\section{Conclusion}
\label{sec:conclusion}
In this paper, we present a novel large scale multimodal dataset of pedestrians at complex urban intersections with a rich set of 2D/3D annotations.
The \textit{PedX} dataset provides a platform for understanding pedestrian behaviors at intersections with real life challenges such as large occlusions, pedestrians walking in group and carrying possessions.
This dataset can be used to solve 3D human pose estimation, pedestrian detection and tracking in-the-wild, and to further expand to new problems. Of particular interest is the negotiation of priority at unsignaled intersections and expanding our prior work in forecasting pedestrian trajectories~\cite{jacobs2017real} to include body pose as a prior in the motion model.

\bibliographystyle{IEEEtran}
\bibliography{main}


\addtolength{\textheight}{-12cm}   




\end{document}